\documentclass[letterpaper]{article} 
\usepackage{aaai24}
\usepackage{times}  
\usepackage{helvet}  
\usepackage{courier}  
\usepackage[hyphens]{url}  
\usepackage{graphicx} 
\usepackage{natbib}  
\usepackage{caption} 
\usepackage{algorithm}
\usepackage{bm}
\usepackage{algpseudocode}
\usepackage{mathtools}
\usepackage{amssymb}
\usepackage{multirow}
\usepackage{multicol}
\usepackage{booktabs}
\usepackage{hyperref}
\usepackage{cleveref}
\usepackage{color}

\usepackage{graphicx}
\usepackage{float} 
\usepackage{adjustbox}
\usepackage{newfloat}
\usepackage{listings}
\usepackage{pifont}
\usepackage{subfig}

\newcommand{\xmark}{\ding{55}}

\newcommand{\yx}[1]{\textcolor{blue}{#1}}
\newcommand{\et}[2]{${#1}^{\pm{#2}}$}
\DeclarePairedDelimiterX{\infdivx}[2]{(}{)}{%
  #1\;\delimsize|\delimsize|\;#2%
}

\usepackage{newfloat}
\usepackage{listings}
\DeclareCaptionStyle{ruled}{labelfont=normalfont,labelsep=colon,strut=off} 
\floatstyle{ruled}
\newfloat{listing}{tb}{lst}{}
\floatname{listing}{Listing}
\title{DiverseMotion: Towards Diverse Human Motion Generation via\\ Discrete Diffusion}
\author{
    Yunhong Lou\textsuperscript{\rm 1},
    Linchao Zhu\textsuperscript{\rm 1},
    Yaxiong Wang\textsuperscript{\rm 2},
    Xiaohan Wang\textsuperscript{\rm 1},
    Yi Yang\textsuperscript{\rm 1},
}
\affiliations{
    \textsuperscript{\rm 1} Zhejiang University\\
    \textsuperscript{\rm 2} Hefei University of Technology\\
}

\usepackage{bibentry}

\begin{document}

\maketitle

\begin{abstract}
We present DiverseMotion, a new approach for synthesizing high-quality human motions conditioned on textual descriptions while preserving motion diversity. 
Despite the recent significant process in text-based human motion generation,
existing methods often prioritize fitting training motions at the expense of action diversity. Consequently, striking a balance between motion quality and diversity remains an unresolved challenge. 
This problem is compounded by two key factors: 1) the lack of diversity in motion-caption pairs in existing benchmarks and 2) the unilateral and biased semantic understanding of the text prompt, focusing primarily on the verb component while neglecting the nuanced distinctions  indicated by other words.
In response to the first issue, we construct a large-scale \textbf{W}ild \textbf{M}otion-\textbf{C}aption  dataset (WMC) to extend the restricted action boundary of existing well-annotated datasets, enabling the learning of diverse motions through a more extensive range of actions. 
To this end, a motion BLIP is trained upon a pretrained vision-language model, then we automatically generate diverse motion captions for the collected motion sequences. As a result, we finally build a dataset comprising 8,888 motions coupled with 141$k$ texts.
To comprehensively understand the text command, we propose a \textbf{H}ierarchical \textbf{S}emantic \textbf{A}ggregation (HSA) module to capture the fine-grained semantics.
Finally,
we involve the above two designs into an effective Motion Discrete Diffusion (MDD) framework  to strike a balance between motion quality and diversity. 
Extensive experiments on HumanML3D and KIT-ML show that our DiverseMotion achieves the state-of-the-art motion quality and competitive motion diversity. Dataset, code, and pretrained models will be released to reproduce all of our results.
Source code will be available on the project page: \url{https://github.com/axdfhj/MDD}

\end{abstract}
\begin{figure}[tb] 
	\centering 
	\includegraphics[width=0.9\linewidth]{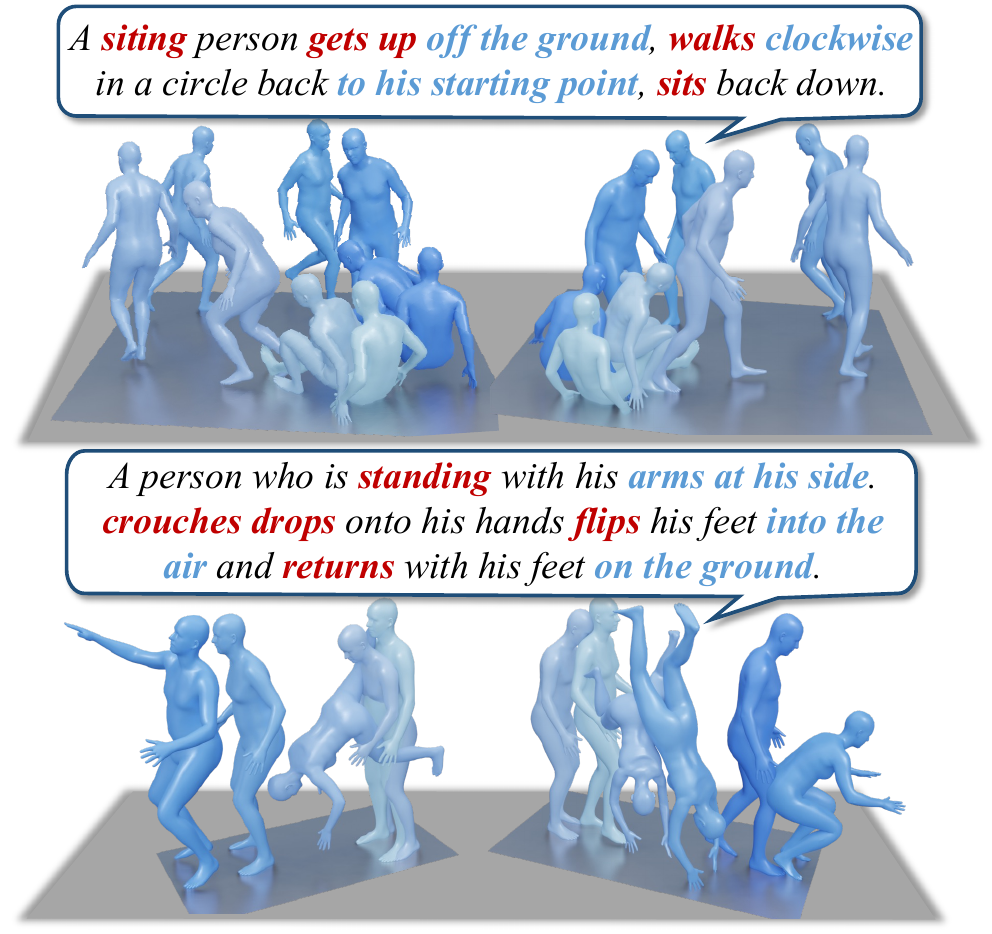} 
	\caption{Our MDD can generate precise and high-quality motions with the aid of our WMC dataset and HSA text encoder. Darker colors indicate later time.} 
	\label{fig:teaser} 
\end{figure} 

\section{Introduction}
The goal of text to motion generation is to synthesize a motion sequence that matching the textual input.  This task has attracted much attention recently due to its importance in many vision applications such as movie, game production, VR/AR and beyond. 
However, traditional methods rely on either professional animators or professional actors with expensive motion capture equipment \cite{heloir2010exploiting, bradwell2008tutorial}, where the formmer manner is labor-intensive and the latter one constrains the availability and affordability of human motion modeling. As a result, researchers have increasingly turned to utilizing advanced generation models to synthesize realistic motions\cite{guo2022generating}. Using autoencoders (AEs) or variational autoencoders (VAEs), JL2P \cite{ahuja2019language2pose} and Ghosh \textit{et al.} \cite{ghosh2023synthesis} modelled a latent space that integrated text and motion, and inferred text-to-motion by decoding text latent codes with trained motion decoders.

However, existing approaches for motion generation primarily focus on enhancing the quality of the generated motions, while pay rare attention to  the diversity of the motion sequences. We argue that this issue can significantly hinder the practical applicability of motion generation models. Specifically, the lack of diversity in generated motion sequences can lead to a monotonous and unengaging experience in virtual environments, where the same action description frequently corresponds to highly similar motion sequences, leading to a template-like quality in the generated actions.  This would restrict the model's applicability in various domains such as computer animation, virtual reality, and human-computer interaction.  
However, common practices prioritize improved quality at the expense of motion diversity, leading to a prevalent scarcity of motion diversity in current state-of-the-art methods.
In our practice,  we observed that one of important factors for this deficiency is attributed to the limited action patterns within existing motion benchmarks.  Anchored on the scarce action categories, the generated motion will inevitably shrink into a compact subspace with the network learning. Another challenge stems from the one-side comprehension of the text prompt, resulting in the semantic embeddings that are dominated by the action verb while disregarding the finer details and nuances associated with the verb.

We identify the fundamental cause of the inadequate diversity are the constrained range of actions in current benchmarks.
A simple solution would involve gathering a substantial and varied set of motion sequences, along with corresponding caption annotations, to mitigate this issue.
Nevertheless, undertaking this procedure manually proves to be a time-consuming and labor-intensive process. Given this consideration, our research aims to automate the construction of a motion-text dataset, thereby streamlining the process and enabling efficient diversity enhancement. It should be noted that numerous motion datasets like NTU RGB+D \cite{shahroudy2016ntu}, Human3.6M \cite{h36m_pami}, GTA-Human \cite{cai2021playing} offer a wealth of abundant and diverse high-quality motion sequences, but lack paired captions. In parallel, recent advancements in cross-modal generation models provide the capability to generate diverse captions for motion sequences. Leveraging this progress, we can harvest a paired and diverse motion-caption dataset. To give reliable captions, we first tune a motion BLIP from a pretrained image-to-text generation model with motion-text pairs sourced from existing datasets \cite{guo2022generating,Plappert_2016}. With the aid of the tuned captioner, we assign textual descriptions to the collected motions.  Notably, the rich words in the pretrained cross-modal model will enable us to generate diverse descriptions.
Finally, we construct a large-scale loosely paired motion-caption dataset, consisting of 8,888 motion clips with 141$k$ textual descriptions, and each motion is associated with 15 sentences on average (See Table ~\ref{wmc}).

Another significant observation derived from our practical is that the current motion generation models often disregard adjunct words associated with the verb. Consequently, prompts expressing the same verb often generate highly similar motion sequences. The essential reason behind this observation is that the text encoder fails to capture the fine-grained semantics in the prompt. To remedy this issue, we propose a Hierarchical Semantic Aggregation (HSA) module, where we adaptive fuse the multi-level semantics of the text encoder, aiming to use the low-level embedding to adaptively supplement the missed details in the high-level semantic representation. In this fashion, we can harvest a comprehensive features of the sentence,  better controlling over the motion sequences and ensuring the inclusion of all details specified in the text command.

To well exploit the potential of the WMC dataset and HSA module, we develop a Motion Discrete Diffusion (MDD) framework. MDD embraces a latent diffusion-style framework with our HSA as the text encoder and learn from the diverse motions in WMC to achieve an optimal balance between motion quality and diversity. Specially, 
motion VQ-VAE encoder discretizes motion as tokens, followed by the application of discrete diffusion on these tokens and the text condition to generate a motion sequence within the latent space. Finally, the VQ-VAE decoder converts the latent motion back into visual motion sequences. During the training phase, we incorporate the WMC dataset directly as auxiliary data instead of using it solely for pretraining purposes. This strategy ensures that the learning process always operates within a wider motion and text space. In summary, the main contributions of this work can be succinctly enumerated as follows:
\begin{table}[t]
\small
\begin{tabular}{c|c|c|c|c}
\toprule
Dataset   & texts    & motions & Duration & Automatic \\ \midrule
HumanML3D & 44,970   & \textbf{14,616}  & 28.59h   & \xmark             \\ \ 
KIT-ML    & 6,278   & 3,911   & 10.33h   &  \xmark             \\ \midrule
WMC       & \textbf{141,010} & 8,888   & \textbf{57.97h}   &  \checkmark             \\ \bottomrule
\end{tabular}
\caption{Comparison of the WMC dataset with HumanML3D~\cite{guo2022generating} and KIT-ML \cite{Plappert_2016}. Our dataset comprises an extensive volume of motion data coupled with textually enriched description providing extensive semantic diversity.}
\label{wmc}
\end{table}

\begin{itemize}
    \item \textbf{A new Motion-Caption dataset} named Wild Motion-Caption (WMC) is constructed, which comprise 141k motion-caption pairs.

    \item \textbf{A quality-diversity balanced model} (MMD) is designed for text-to-motion generation, where we effectively utilize the WMC dataset and HSA module to balance the quality and diversity. 

    \item \textbf{Competitive performance.} Our MMD achieves state-of-the-art performance with 0.072 FID and 2.941 MM-Dist on HumanML3D \cite{guo2022generating} dataset.
\end{itemize}

\begin{figure*}[t] 
	\centering 
	\includegraphics[width=1\linewidth]{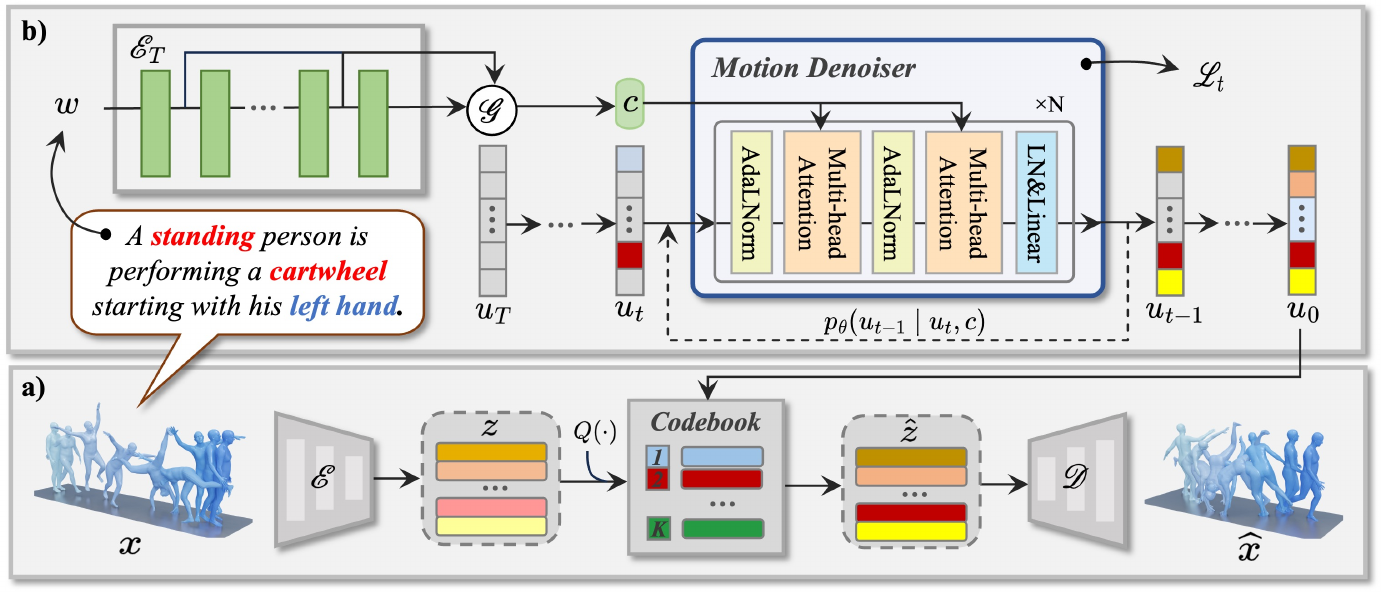} 
	\caption{\textbf{Overview of the method.} MDD contains two training stage. a) trains an encoder $\mathcal{E}$, a decoder $\mathcal{D}$ and a codebook $C$ by reconstructing motions. b) trains a motion denoiser $p_\theta(\bm{u}_{t-1}|\bm{u}_t, c)$ to reverse a Markov chain conditioned on text $w$. In the inference stage, the motion denoiser generate motion tokens $u_0$ from fully masked tokens $u_T$ and then we decode $u_0$ to get natural human motion with decoder $\mathcal{D}$.}
	\label{framework} 
\end{figure*}

\section{Related Work}

\noindent\textbf{Diffusion Probabilistic Models} \cite{sohldickstein2015deep,song2021scorebased} have recently emerged as a powerful generative modeling approach for high-fidelity image synthesis \cite{ho2020denoising,ho2021cascaded,dhariwal2021diffusion}. 
By gradually corrupting a data sample with Gaussian noise and then training a model to reverse this process, diffusion models can capture complex data distributions with two Markov chains while avoiding common failure modes like mode collapse. 
Following early work on image generation, recent efforts \cite{austin2023structured,dhariwal2021diffusion} have explored conditioning diffusion models on textual prompts to perform controllable image synthesis from text descriptions. Additionally, a classifier-free \cite{ho2022classifierfree} method is proposed to randomly drop out the condition to balance generation diversity and generation quality~\cite{nichol2022glide}.
Recently, another discrete version of diffusion probabilistic model \cite{austin2023structured,sohldickstein2015deep,hoogeboom2021argmax} has attracted much attention with more controllable corruption and denoising processes \cite{austin2023structured}, better input-output consistency \cite{zhu2023discrete} and excellent generative diversity \cite{gu2022vector}, which has been verified to have promising performance in multiple conditional generation tasks such as text-to-image \cite{gu2022vector}, dance-to-music \cite{zhu2023discrete}, category-to-layout \cite{inoue2023layoutdm}.

\noindent\textbf{Text-driven Human Motion Generation.} Human motion modeling has a long research history \cite{10.1093/oso/9780195073591.001.0001}. 
Recent years, the field of text-driven human motion generation has attracted the attention of researchers.
Early research focused on modeling a joint latent space of motion and language using Autoencoder (AE) or Variational Autoencoder (VAE). 
JL2P \cite{ahuja2019language2pose} implements the construction of joint latent spaces with a curriculum learning approach. 
Ghosh \textit{et al.} \cite{ghosh2023synthesis} uses a hierarchical approach to model human upper and lower limb movements separately in the hidden space. 
Guo \textit{et al.} \cite{guo2022generating} construct a currently largest public motion-text dataset and propose an auto-regressive VAE to address the problem. 
Motiondiffuse \cite{zhang2022motiondiffuse} and MDM \cite{tevet2022human} apply diffusion model to text-driven motion generation tasks during the same period, significantly increase the performance. MLD \cite{chen2023executing} models the latent space of a VAE with diffusion model. T2M-GPT \cite{zhang2023t2mgpt} tokenizes the raw motion into discrete tokens and generate it with GPT architecture in auto-regressive manner.


\section{Preliminary}
Before diving into our method, we first introduce the foundation architectures of our Motion Discrete Diffusion (MDD): Motion VQ-VAE and Discrete Diffusion model.

\subsection{Motion VQ-VAE}

Motion VQ-VAE \cite{oord2018neural,razavi2019generating, zhang2023t2mgpt} targets to effectively compress and discrete motion, 
which consists an encoder $ \mathcal{E} $, a decoder $ \mathcal{D} $ and a motion codebook $ \mathcal{\bm{C}} = {\{\bm{c}_k\}}_{k=1}^{K} \in \mathbb{R}^{K \times d} $ containing a finite number of motion features
, where $ K $ is the size of the motion codebook and $ d $ is the dimension of embedding vectors.
Presented with a $N$ frame motion sequence $ \bm{m} \in \mathbb{R}^{N \times d} $, the estimated latent embedding can be obtained by computing $ \bm{z} = \mathcal{E}(\bm{m}) = {\{\bm{z}^i\}}_{i=1}^{n} \in \mathbb{R}^{n \times d} $, where the downsampling rate 
$\eta$ is denoted by $ n/N $. After that, {a temporal quantizer} $ Q(\cdot) $ is employed to find the nearest embedding $\bm{z}_q$ in the codebook $\mathcal{\bm{C}}$:
\begin{equation}
	\bm{z}_q^i = Q\left( \bm{z}^i \right) =\left( \underset{\bm{c}_k\in \mathcal{\bm{C}}}\arg\min\|\bm{z}^i - \bm{c}_k\|_2 \right) \in \mathbb{R}^{1 \times d}
\end{equation}
The objective of training Motion VQ-VAE is  to reconstruct $\bm{m}$ into $\tilde{\bm{m}}=\mathcal{D}(\bm{z}_q)$:
\begin{equation}
    \mathcal{L}_{VQ} = \| \bm{m} - \tilde{\bm{m}} \|_1 + \underbrace{||\bm{z} - \mathit{sg}[\bm{z}_q]||_2}_{\mathcal{L}_{embed}} + \beta \underbrace{||\mathit{sg}[\bm{z}] - \bm{z}_q||_2}_{\mathcal{L}_{commit}}
\end{equation}
where $sg[\cdot]$ refers to the stop-gradient operation and $\beta$ refers to a hyper-parameter. 
We update the codebook with exponential moving averages (EMA) and codebook reset method which reinitializes parameters of inactive code with active ones during training \cite{williams2020hierarchical, zhang2023t2mgpt}, both of which has been proven to be very effective in previous works\cite{zhang2023t2mgpt,razavi2019generating}.

\subsection{Discrete Diffusion for Motion Generation.}
T2M-GPT \cite{zhang2023t2mgpt} generates code indexes with an auto-regressive fashion which has the inherent flow of the unidirectional bias and lead to error accumulation \cite{sheng2019woman}, resulting in insufficient generative diversity. Additionally, T2M-GPT is not capable of controlling the duration of the motion. To avoid these issues, we abandon the T2M-GPT-based framework but carefully design a discrete diffusion process for motion generation tasks, thereby maintaining the advantages of diffusion model for continuous spaces, such as high fidelity and substantial diversity.
With the aid of the trained Motion VQ-VAE, we can effectively conduct discrete diffusion in the latent space. The process begins by encoding the text input $ \bm{w} = \{w^i\}_{i=1}^{L} $ using text encoder and sampling a fully masked motion tokens.

\noindent\textbf{Conditional Discrete Diffusion.} 
During the training stage of Discrete Diffusion, given a sequence of human motion $\bm{m}^{1:N}$ paired with a textual description $\bm{w}^{1:L}$, we first tokenize $\bm{m}$ to discrete motion tokens $\bm{u}_0^{1:n}$ with trained Motion VQ-VAE encoder $\mathcal{E}$ and encode $\bm{w}$ to embedding $c$ with text encoder $\mathcal{E}_T$
. In addition, a time step $t \sim \mathcal{U}(1,T)$ is sampled to determine the current training step.

\textit{Forward diffusion process} $ q(\bm{u}_t|\bm{u}_{t-1})$  is defined to corrupt $\bm{u}_{t-1}$ by resampling some tokens of it uniformly with a pre-scheduled chance $\beta_t$, where $q(\bm{u}_T) \sim \mathcal{U}(1,K)$. Suppose we are given a deterministic motion sequence $\bm{u}_{t-1} = \{u^i_{t-1}\}_{i=1}^{n}$ at time step $t-1$, corrupted function $q(u_t^i|u_{t-1}^i)$ can be presented as:
\begin{equation}
    q(u_t^i = k|u_{t-1}^i) = \label{eq:trans}
    \begin{cases}
        \beta_t / K & k \neq u^i_{t-1} \\
        1 - \beta_t(K-1)/K & k = u^i_{t-1}
    \end{cases}
\end{equation}
Due to the Markovian nature of the forward diffusion process, we derive the probability distribution $q(\bm{u}_t|\bm{u}_0)=\mathcal{\bm{P}}(\bm{u}_t|\overline{\alpha}_t\bm{u}_0+(1-\overline{\alpha}_t)/K)$ for any step $t$ given $\bm{u}_0$ 
, where $\alpha_i = 1 - \beta_i $, $ \overline{\alpha}_t = \prod_{i=1}^t \alpha_i$. $\mathcal{\bm{P}}$ denotes how to express the distribution of random variable before $|$ with given parameters after $|$.

\textit{Reverse diffusion process} works by training a denoising network $p_\theta(\bm{u}_{t-1}|\bm{u}_t, c)$ conditioned on text embedding $c$ and time step $t$ to estimate posterior distribution $q(\bm{u}_{t-1}|\bm{u}_t,\bm{u}_0)$. Our optimization objective $\mathcal{L}$ is mathematically consistent with the DDPM:
\begin{equation}
\begin{split}
\mathcal{L}_t = 
\begin{cases}
    -\log p_\theta(\bm{u}_0|\bm{u}_1, c, t) & t = 0 \\
    D_{\text{KL}}(q\left(\bm{u}_{t-1} | \bm{u}_t, \bm{u}_0\right) \| p_\theta\left(\bm{u}_{t-1} | \bm{u}_t, c\right)) & t>0 \\
    D_{\text{KL}}(q\left(\bm{u}_{T} | \bm{u}_0\right) \| p\left(\bm{u}_{T}\right)) & t=T  
\end{cases}
\end{split}
\end{equation}

\textit{Masked Language Diffusion.} As WMC dataset is characterized by extensive motion but relatively loose caption compared to manually labeled data, we apply mask with different probabilities $\sigma$ to text from different sources, which is effective in avoiding semantic collapse. In this way, We set $w=\emptyset$ to learn unimodal denoising and the ability of unconditional generation. To simplify the denoising, we introduced the mask-and-replace \cite{gu2022vector} to explicitly distinguish between clean and noisy tokens of $u_{t-1}$. Specifically, an additional mask token denoted $K+1$ is introduced, which won't change during the forward process once the token becomes it. Corrupted function of Eq.\ref{eq:trans} for non-mask token $(u^i_{t-1} \neq K+1)$ becomes:
\begin{equation}
    q(u_t^i = k|u_{t-1}^i) =
    \begin{cases}
        \beta_t / K & k \neq u^i_{t-1} \\
        \alpha_t + \beta_t / K & k = u^i_{t-1} \\
        1-\alpha_t-\beta_t & k=K+1
    \end{cases}
\end{equation}
In addition, $\bm{u}_T$ set to be fully masked, also serving as the starting for reverse process.

\section{Methodology}


Figure~\ref{framework} illustrates the framework of our MDD. Motion VQ-VAE encoder
is responsible to compress the raw motion, such that the subsequent discrete motion can be effectively performed in the low-dimensional space.  Specially, 
the motion VQ-VAE encoder discretizes motion as tokens, and the representation of text prompt is encoded via our HSA module. Both serve as the input for the discrete diffusion  to generate a motion sequence within the latent space. Finally, the VQ-VAE decoder converts the latent motion back into visual motion sequences.
During the training phase,  we directly include the WMC dataset as auxiliary training data rather than solely for pretraining, targeting to maintain an expanded motion and text space within the learning process, which is effective to  prevent the motion mapping from shrinking into a narrow space. 

\subsection{Wild Motion-Caption Dataset}
Augmenting motion-caption pairs helps to facilitate diverse motion generation.
First, a wealth of single-modality motion sequences are readily available like Human3.6M \cite{h36m_pami}, constituting valuable and high-quality motion resources for network learning. Secondly, the remarkable progress in vision-to-text generation models enables the reliable captioning of motions, thereby allowing for the compilation of high-quality motion-caption pairs that form the foundation of our WMC dataset. The construction of our dataset is composed of the following stages:

\noindent\textbf{Motion Collection.} Our motion data is sourced from three existing public motion datasets, RICH \cite{Huang:CVPR:2022}, Human3.6M \cite{h36m_pami} and FLAG3D \cite{tang2023flag3d}. Among them, Human3.6M and FLAG3D are collected through high-precision MoCap equipment in a laboratory environment. RICH is an accurately annotated motion dataset through markerless motion capture method in natural environment.

\noindent\textbf{Motion Preprocess.} To bridge the gap between different datasets, we align motion data from various sources to the widely-used and large-scale HumanML3D dataset\cite{guo2022generating}. We obtain the first 22 key points of SMPL \cite{bogo2016smpl,pavlakos2019expressive} skeleton and align them to the common human skeleton template used in HumanML3D. All the motions are downsampled to 20 FPS and cropped to be within 196 frames. We shift starting points to the origin of the world coordinate system and rotate motion in XZ-plane to orient the initial state facing the Z+ axis.

\noindent\textbf{Motion BLIP Training.} We utilize the off-the-shelf multimodal generation model BLIP-2 \cite{li2023blip2} as our captioner, which has the ability of open-ended visual content understanding and language generation. BLIP-2 comprises three main components: a vision transformer (VIT) that extracts visual features from a single image, a language model that generates text content, and a Q-former that connects the visual and linguistic domains. To generate texts that focus on the motion of the presented images rather than the caption style in the dataset used for BLIP-2 model training, we tune the model with pair data from the existing motion-language dataset. First, we render the motion data into a sequence of images. To better utilize BLIP-2 for captioning temporal-continuous image sequences, we concatenate the image tokens generated from VIT and apply a learnable temporal embedding to each image. The parameters of the language model are frozen, while the parameters of the VIT and Q-former are trained to align with the stylistic characteristics of rendered images. The objective of the tuning is as follows:
\begin{table*}[!t]
    \centering
    \footnotesize
    \adjustbox{max width=\linewidth}{

    \begin{tabular}{l c c c c c c c}
    \toprule
    \multirow{2}{*}{Methods}  & \multicolumn{3}{c}{R-Precision $\uparrow$} & \multirow{2}{*}{FID $\downarrow$} & \multirow{2}{*}{MM-Dist $\downarrow$} & \multirow{2}{*}{Diversity $\rightarrow$} & \multirow{2}{*}{MModality $\uparrow$}\\

    \cmidrule{2-4}
    ~ & Top-1 & Top-2 & Top-3 \\

    \midrule

        \textbf{Real motion} & \et{0.511}{.003} & \et{0.703}{.003} & \et{0.797}{.002} & \et{0.002}{.000} & \et{2.974}{.008} & \et{9.503}{.065} & -  \\
        VQ-VAE \small{(w/o)} & \et{0.496}{.002} & \et{0.689}{.002} & \et{0.787}{.002} & \et{0.070}{.001}& \et{3.072}{.009} & \et{9.593}{.079} & -  \\
        VQ-VAE & \et{0.491}{.003} & \et{0.684}{.002} & \et{0.781}{.002} & \et{0.064}{.001}& \et{3.097}{.008} & \et{9.653}{.075} & -  \\
    \midrule






        TM2T & \et{0.424}{.003} & \et{0.618}{.003} & \et{0.729}{.002} & \et{1.501}{.017} & \et{3.467}{.011} & \et{8.589}{.076} & \et{\underline{2.424}}{.093}  \\

        Guo \textit{et al.} & \et{0.455}{.003} & \et{0.636}{.003} & \et{0.736}{.002} & \et{1.087}{.021} & \et{3.347}{.008} & \et{9.175}{.083} & \et{2.219}{.074}  \\

        MotionDiffuse  & \et{{0.491}}{.001} & \et{{0.681}}{.001} & \et{{0.782}}{.001} & \et{0.630}{.001} & \et{{3.113}}{.001} & \et{9.410}{.049} & \et{1.553}{.042}  \\

        MDM & \et{0.320}{.005} & \et{0.498}{.004} & \et{0.611}{.007} & \et{0.544}{.044} & \et{5.566}{.027} & \et{\underline{9.559}}{.086} & \et{\textbf{2.799}}{.072}  \\

        MLD & \et{0.481}{.003} & \et{0.673}{.003} & \et{0.772}{.002} & \et{0.473}{.013} & \et{3.196}{.010} & \et{{9.724}}{.082} & \et{2.413}{.079}  \\

        T2M-GPT & \et{{0.491}}{.003} & \et{0.680}{.003} & \et{0.775}{.002} & \et{{0.116}}{.004} & \et{3.118}{.011} & \et{{9.761}}{.081} &  \et{1.856}{.011} \\
    \midrule
        Ours $s=1$ & \et{\underline{0.496}}{.004} & \et{\underline{0.687}}{.004} & \et{\underline{0.783}}{.003} & \et{\textbf{0.070}}{.004} & \et{\underline{3.063}}{.011} & \et{\textbf{9.551}}{.068} & \et{2.062}{.079} \\
        Ours $s=2$ & \et{\textbf{0.515}}{.003} & \et{\textbf{0.706}}{.002} & \et{\textbf{0.802}}{.002} & \et{\underline{0.072}}{.004} & \et{\textbf{2.941}}{.007} & \et{{9.683}}{.102} & \et{1.869}{.089} \\
    \bottomrule
    \end{tabular}
    }
    \caption{\textbf{Comparison with the state-of-the-art methods on HumanML3D~\cite{guo2022generating} test set.} We compute suggested metrics following Guo \textit{et al.}~\cite{guo2022generating}. For each metric, we repeat the evaluation 20 times (except \textit{MModality} runs 5 times) and report the average with a 95\% confidence interval. $\rightarrow$ indicates that the closer to the real data, the better. \textbf{Bold} and \underline{underline} indicate the best and the second best result. w/o denotes training without WMC. $s$ denotes the classifier-free scale.}
    \label{Table 1}

\end{table*}
\begin{equation}
    \mathcal{L}_{\theta}=-\frac{1}{|\boldsymbol{t}|}\sum_{i=1}^{|\boldsymbol{t}|}\log \mathcal{\bm{H}}_{\theta}\left(\boldsymbol{t}_{i}\mid\boldsymbol{t}_{<i}\right)
\end{equation}

where $\mathcal{\bm{H}}$ means the network of our captioner with trainable parameters $\theta$, $\bm{t}_i$ stands for the ${i}^{th}$ token of input.

\noindent\textbf{Diverse Caption Generation.} We employ nucleus sampling \cite{holtzman2020curious} to generate several diverse high-quality captions for each motion sequence. Due to lack of controllable generation capabilities of the captioner, some generated captions fail to meet the requisite quality standards for accurately portraying the associated action. We use MM-Dist \cite{guo2022generating}, a metric representing text-motion consistency, to automate the evaluation of each generated caption. Caption with an MM-Dist value exceeding a predetermined threshold of $\tau$ is deemed unreasonable and deleted.
\begin{table*}[t]
    \centering
    \footnotesize
    \scalebox{1}{

    \begin{tabular}{l c c c c c c c}
    \toprule
    \multirow{2}{*}{Methods}  & \multicolumn{3}{c}{R-Precision $\uparrow$} & \multirow{2}{*}{FID $\downarrow$} & \multirow{2}{*}{MM-Dist $\downarrow$} & \multirow{2}{*}{Diversity $\rightarrow$} & \multirow{2}{*}{MModality $\uparrow$}\\

    \cmidrule{2-4}
    ~ & Top-1 & Top-2 & Top-3 \\

    \midrule

        \textbf{Real motion} & \et{0.424}{.005} & \et{0.649}{.006} & \et{0.779}{.006} & \et{0.031}{.004} & \et{2.788}{.012} & \et{11.080}{.097} & -  \\
        VQ-VAE & \et{0.402}{.006} & \et{0.614}{.007} & \et{0.745}{.008} & \et{0.374}{.009}& \et{2.942}{.015} & \et{10.955}{.114} & -  \\
    \midrule






        TM2T  & \et{0.280}{.005} & \et{0.463}{.006} & \et{0.587}{.005} & \et{3.599}{.153} & \et{4.591}{.026} & \et{9.473}{.117} & \et{\textbf{3.292}}{.081}  \\

        Guo \textit{et al.} & \et{0.361}{.006} & \et{0.559}{.007} & \et{0.681}{.007} & \et{3.022}{.107} & \et{3.488}{.028} & \et{10.72}{.145} & \et{2.052}{.107}  \\

        MDM & \et{0.164}{.004} & \et{0.291}{.004} & \et{0.396}{.004} & \et{0.497}{.021} & \et{9.191}{.022} & \et{10.847}{.109} & \et{1.907}{.214}  \\

        MotionDiffuse  & \et{\textbf{0.417}}{.004} & \et{0.621}{.004} & \et{0.739}{.004} & \et{1.954}{.062} & \et{\underline{2.958}}{.005} & \et{\textbf{11.100}}{.143} & \et{0.730}{.013}  \\

        MLD & \et{0.390}{.008} & \et{0.609}{.008} & \et{0.734}{.007} & \et{\textbf{0.404}}{.027} & \et{3.204}{.027} & \et{{10.800}}{.117} & \et{\underline{2.192}}{.071}  \\

        T2M-GPT & \et{\underline{0.416}}{.006}  & \et{\underline{0.627}}{.006} & \et{\underline{0.745}}{.006} & \et{0.514}{.029} & \et{3.007}{.023} & \et{\underline{10.921}}{.108} & \et{1.570}{.039} \\
    \midrule
        Ours & \et{\underline{0.416}}{.005} & \et{\textbf{0.637}}{.008} & \et{\textbf{0.760}}{.011} & \et{\underline{0.468}}{.098} & \et{\textbf{2.892}}{.041} & \et{{10.873}}{.101} & \et{2.062}{.079} \\
    \bottomrule
    \end{tabular}
    }
    \caption{\textbf{Comparison with the state-of-the-art methods on KIT-ML~\cite{Plappert_2016} test set.} We compute suggested metrics following Guo \textit{et al.}~\cite{guo2022generating}. For each metric, we repeat the evaluation 20 times (except \textit{MModality} runs 5 times) and report the average with 95\% confidence interval. }
    \label{Table 2}

\end{table*}

\noindent\textbf{WMC Dataset.} Following the above steps, a large-scale dataset of loosely motion-text is built. As shown in \cref{Table 1}, WMC comprise 141$k$ texts with 8.9$k$ motion clips.

\subsection{Hierarchy Semantic Aggregation.} 
We use a frozen CLIP model as our text encoder $\mathcal{E}_T$. However, single-layer output of CLIP cannot capture the semantic information at different levels of granularity \cite{ahuja2019language2pose}, which may lead to overemphasizing some semantics, such as a specific verb, and ignoring  some important motions attributes, such as speed, direction, or another concurrent action. Therefore, we propose a simple and effective Hierarchy Semantic Aggregation (HSA) to integrate text features of different granularities as a way to cope with complex motion descriptions. Precisely, given a set $\mathcal{\bm{S}}$ selected in advance, the output $c$ is calculated as:

\begin{equation}
    c=\sum_{i \in \mathcal{\bm{S}}}  a_i \mathcal{F}_i(\mathcal{E}_T^i(\bm{w})),
\end{equation}
where $a_i$, $\mathcal{F}_i(\cdot)$ and $\mathcal{E}_T^i(\cdot)$ denote a learnable weight, an MLP layer and output feature of $i^{th}$ layer of original CLIP model separately. Equipped with supplementary textual information provided by low-level features, more precise conditional guidance in denoising stage leads to more accurate and richer motion details.

\subsection{Balanced Inference with Hybrid Guidance.}
Inspired by the classifier-free method \cite{ho2022classifierfree}, we set an adjustable hyperparameter $s$ to scale the distributions of conditional and unconditional manners:
\begin{equation}
    \widetilde{\mathcal{P}}(\bm{u_{t-1}}|\bm{u_{t}},c) = (1+s){\mathcal{P}}(\bm{u_{t-1}}|\bm{u_{t}},c)-s{\mathcal{P}}(\bm{u_{t-1}}|\bm{u_{t}})
\end{equation}
which we found to be effective in regulating the diversity and the semantic consistency of the motion-text. Additionally, we replace all negative values 
with 0 to enable gumbel soft-max \cite{jang2017categorical}. The final predicted motion tokens are fed into the decoder of Motion VQ-VAE to produce the motion sequence.

\section{Experiments}

\subsection{Datasets and Evaluation Metrics}

\noindent\textbf{HumanML3D}~\cite{guo2022generating} is currently the most largest 3D motion-language dataset in terms of both motion and textual descriptions, which comprises 14,616 realistic human motions along with 44,970 text descriptions, with average length of 12 words per description. 
All the motions are well standardized to a standard human skeleton template, and downsampled to 20 FPS.

\noindent\textbf{KIT-ML}~\cite{Plappert_2016} contains 3,911 human motion sequences and each motion is annotated with 1-4 natural language descriptions. In total, the dataset contains 6,278 textual descriptions with an average length of 8 words.




\noindent\textbf{Evaluation Metrics.} 
Following the previous practice \cite{guo2022generating}, we report following metrics for performance evaluation: \textbf{FID} (Frechet Inception Distance),  \textbf{MM-Dist}(Multi-modal Distance), and \textbf{R-Precision} at rank 1,2,3 for the  evaluation of motion quality and semantic consistency,  \textbf{Diversity} and  \textbf{MModality}(Multimodality) are used to access the motion diversity.

\subsection{Implementation Details} 



For constructing WMC, we adopt BLIP-2 VIT-G OPT-2.7B as the fundational captioner. The tuning process spans 23 epochs, utilizing HumanML3D's training set with a learning rate of 1e-5. Our VQ-VAE's model structure and hyperparameter design are aligned with those presented in \cite{zhang2023t2mgpt}. The training process for VQ-VAE spans 1800 epochs, employing a batch size of 512. Regarding HSA, we adopt CLIP VIT-B/32 as a fundational text encoder with a designated layer set $S=\{7, 9, 11, 12\}$. Our motion denoiser comprises 17 transformer layers. The denoiser undergoes training for 975 epochs with batch size 256. Throughout our training procedures, an optimizer with a weight deca of 4.5e-2 is employed. While VQ-VAE is trained on 1 Tesla A100 for 5 hours, the captioner and denoiser are trained on 4 Tesla A100 for 16 hours and 17 hours separately.


\subsection{Comparison to State-of-the-art Approaches}

\textbf{Quantitative results.} \Cref{Table 1} and \Cref{Table 2} compare our results with current state-of-the-art approaches \cite{guo2022tm2t,guo2022generating,tevet2022human,zhang2022motiondiffuse, chen2023executing,zhang2023t2mgpt} with metrics suggested by \cite{guo2022generating}. We compute the average of the metrics over 20 trials within a 95\% confidence interval to present our experimental findings, except for MModality, which is replicated five times. Experimental results on HumanML3D, as illustrated in Table \Cref{Table 1}, reveal a notable superiority of our method over previous methods concerning FID and  motion-text consistency related metrics (MM-Dist and R-Precision). Our motion-text consistency metrics can even exceed that of real motion by enhancing the effect of the textual condition during inference stage. Experimental results on KIT-ML \Cref{Table 1} show that when training on a small amount of data, our method also has decent performance with best text-motion consistency and second best FID. Compared to T2M-GPT which is built to generate discrete motion tokens using the GPT, our method performs better in almost all metrics.

\begin{figure*}[t] 
	\centering 
	\includegraphics[width=1\linewidth]{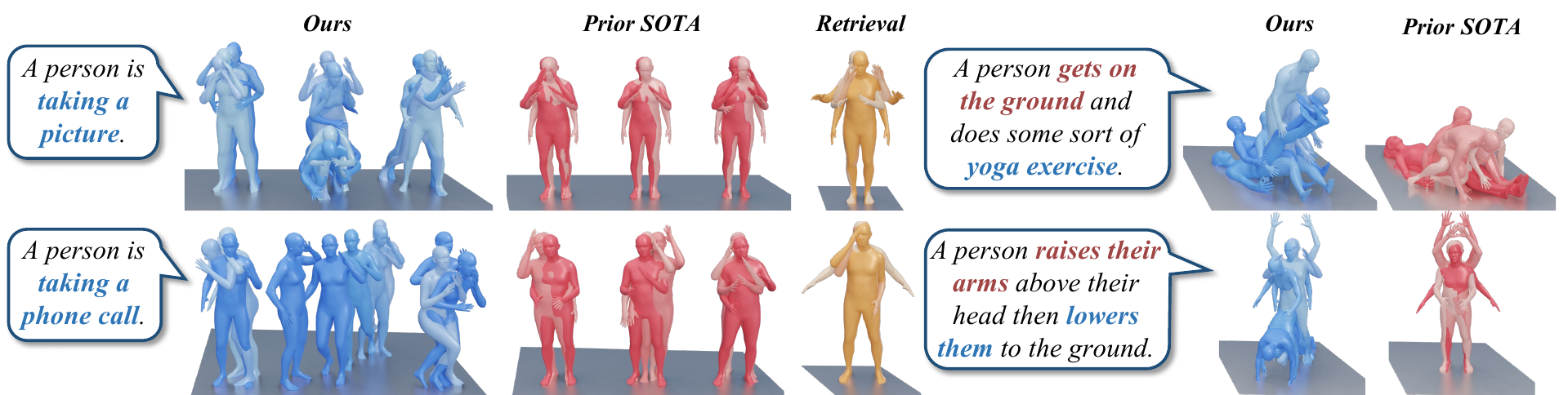} 
	\caption{\textbf{Visual Comparison of the state-of-the-art methods} on text-to-motion generation. The colors from light to dark indicate the progression of time. MDD is presented in blue, the prior SOTA is presented in red, and the retrieval results of HumanML3D are presented in yellow. (\textbf{left}) demonstrates that our approach can produce more plausible results for a given prompt. (\textbf{right}) demonstrates some new motion patterns or combinations learned by our approach.}
	\label{Fig:visual} 
\end{figure*}

\noindent\textbf{Qualitative results.} We present our qualitative results in Figure~\ref{Fig:visual}. Compared to previous SOTAs, the generative diversity of our model is much better than existing models. We present our generative diversity from two perspectives. One is that for the same text, our model can generate a variety of plausible generation results learning from broader motion pattern. In Figure~\ref{Fig:visual} (\textbf{left}), the prior state-of-the-art method generally generates results in a manner similar to the retrieval result \footnote{The retrieval results are obtained by the state-of-the-art text-motion retrieval method TMR \cite{petrovich23tmr}.}, whereas our approach can provide more plausible results. As shown in Figure~\ref{Fig:visual} (\textbf{right}), our approach works as well for some new types of actions or new combinations of simple actions.

%

\begin{figure}[tb] 
	\centering 
        
	\includegraphics[width=1\linewidth]{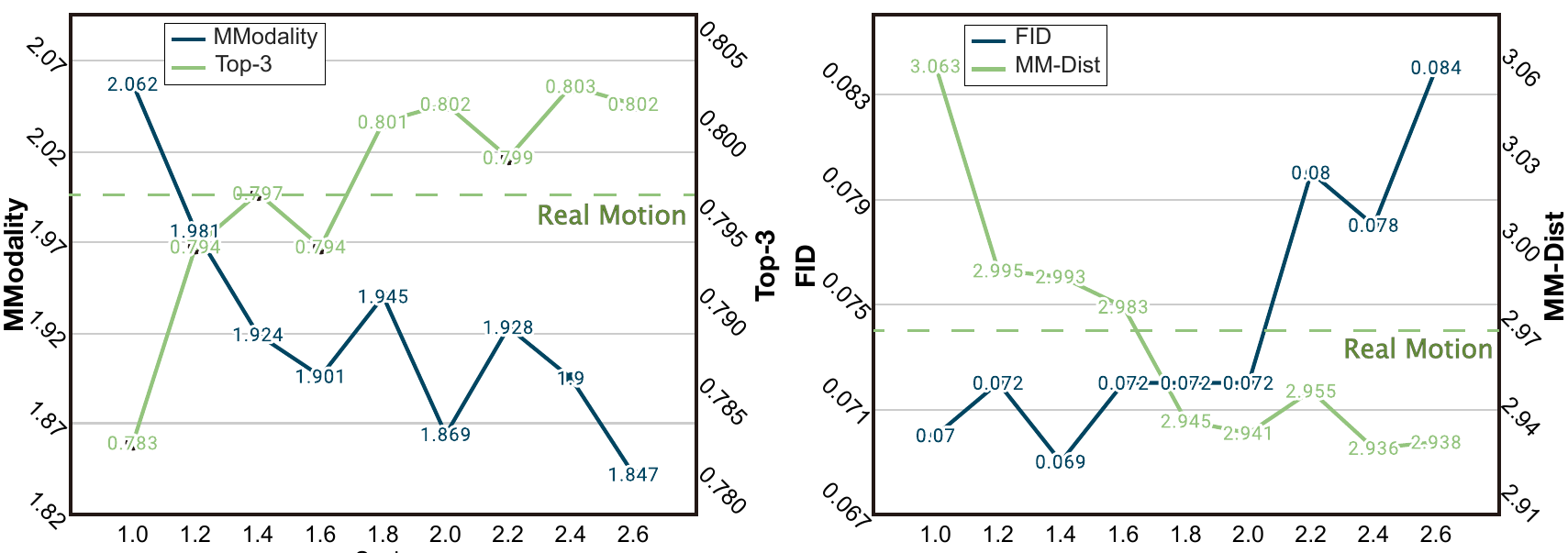} 
	\caption{\textbf{Ablation experiments of the scale $s$.}} 
	\label{fig:scale} 
\end{figure} 

\begin{table}[t]
\centering
\small

\resizebox{1.\linewidth}{!}{
\begin{tabular}{ccccccc}
\toprule\

\textbf{Methods} & \textbf{FID} $\downarrow$ & \textbf{MM-Dist} $\downarrow$ & \textbf{MM-Dist-L} $\downarrow$ & \textbf{Top-1} $\uparrow$ &\textbf{ Top-1-L} $\uparrow$\\
\midrule
Real Mot & 0.002 & 2.974 & 3.073 & 0.511 & 0.496 \\
SOTA & {0.116} & 3.113 & - & 0.491 & - \\
\midrule
Baseline & 0.135 & {3.101} & 3.291 & 0.484 & 0.454 \\
Ours & {\textbf{0.070}} & {\textbf{3.063}} & \textbf{3.234} & {\textbf{0.496}} & \textbf{0.469}\\
\midrule
$\Delta$ & 0.065 & 0.038 & 0.057 & 0.012  & 0.015\\
\bottomrule
\end{tabular}
}
\caption{\textbf{Ablation of HSA module.} MM-Dist-L and Top-1-L denote evaluation results on text no less than 12 words.}
\label{Table 4}
\end{table}


\begin{table}[t]

\adjustbox{max width=\linewidth}{
\begin{tabular}{ccccccc}
\toprule\

\textbf{Step} & \textbf{FID} $\downarrow$ & \textbf{MM-Dist} $\downarrow$ & \textbf{Top-1} $\uparrow$ & \textbf{Top-3} $\uparrow$ & \textbf{MModality} $\uparrow$\\
\midrule
30 & 0.092 & \textbf{3.052} & \underline{0.496} & \textbf{0.784} & 2.211 \\
50 & 0.096 & \underline{3.054} & \textbf{0.498} & \textbf{0.784} & 2.120 \\
80 & 0.087 & 3.075 & 0.492 & 0.777 & \textbf{2.332} \\
100 & \textbf{0.070} & {3.063} & \underline{0.496} & 0.783 & 2.062 \\
120 & \underline{0.075} & 3.084 & 0.493 & 0.780 & \underline{2.271} \\
\bottomrule
\end{tabular}}
\caption{\textbf{Ablation of Diffusion Step.}}
\label{Table:diffusion}
\end{table}

\noindent\textbf{Effects of Hierarchy Semantic Aggregation.} 
\Cref{Table 4} discusses the performance effects of our HSA module. As for the baseline model, we use the word-wise text tokens 
$c \in \mathbb{R}^{77 \times 256}$ from the CLIP model as our textual condition. Compared to the baseline, our approach results in an extremely improved the FID of the generated results by 48\%. It is also clear that our approach can still further improve the consistency of the text with the motions, although the baseline results are already new sota compared to the previous work, which  also verifies the ability of HSA to extract comprehensive semantics. To test the capabilities of our HSA in more complex utterance scenarios, we filter the test set of sentences with less than 12 words, with MM-Dist and Top-1 being denoted MM-Dist-L and Top-1-L respectively. It can be noticed that our method improves larger margins when dealing with complex motion descriptions.

\noindent\textbf{Effects of classifier-free scale.} In Figure~\ref{fig:scale}, we evaluate the performance of our model with different classifier-free scale $s$ in the HumanML3D test set. The results show that this is a very flexible way to regulate the relationship between motion-text consistency and generative diversity in the inference phase. When we amplify the effect of text-condition by increasing $s$, text-motion consistency related metrics can even exceed real motion, but at the same time, it can lead to a decrease in generative diversity. 

\noindent\textbf{Effects of Discrete Diffusion Steps.} We explore the effect of different diffusion steps on the quality of the generated motions. The results are reported in Table~\ref{Table:diffusion}, we can observe that the setting of step 100 is a good configuration on average, too few diffusion steps is insufficient to reach the performance peak, while too large diffusion is hypercorrect.


\section{Conclusion}
In this work, we introduce a novel human motion generation model called Motion Discrete Diffusion (MDD), which seeks to achieve a balance between query-diversity in the generated motions. Our investigation reveals that the current lack of diversity in motion generation can be attributed to two primary factors: inadequate diversity within existing motion benchmarks and a one-sided semantic interpretation of the text commands.
To remedy these issues, we propose a new motion benchmark, termed wild motion-caption (WMC), specifically designed to provide a broader range of motion diversity. Additionally, we present a Hierarchical Semantic Aggregation (HSA) module that aims to capture finer-grained semantics within the text prompts. By integrating the WMC benchmark and HSA module, we establish our MDD framework based on the Motion VQ-VAE and the discrete diffusion, which effectively achieves a balanced combination of motion quality and diversity.
Extensive experimentation conducted on widely-adopted benchmark datasets demonstrates the efficacy of our proposed model, as it successfully attains a favorable trade-off between motion quality and diversity.

\bibliography{ref}
\clearpage
\appendix
\vspace*{1em}{\centering\large\bf%
Appendix
\vspace*{1.5em}}


\section{Motion Representations}
The representation of motion $ \bm{x}^{1:N} $ is distinct in different datasets. For our setting, we align the motions from WMC with the redundant motion representation \cite{Peng_2021,10.1145/3528223.3530178,10.1145/3355089.3356505}
used in HumanML3D. Each pose is defined as a tuple: $ x^i = \{\dot{r}^a, \dot{r}^x, \dot{r}^z, r^y, \mathbf{j}^p, \mathbf{j}^v, \mathbf{j}^r, \mathbf{c}^f\} $, where $ \dot{r}^a, \dot{r}^x, \dot{r}^z \in \mathbb{R} $ are the global root angular velocity and the global root velocity in the X-Z plane; $ r^y \in \mathbb{R} $ represents the height of root. $\mathbf{j}^p \in \mathbb{R}^{3j}, \mathbf{j}^v \in \mathbb{R}^{3j}, \mathbf{j}^r \in \mathbb{R}^{6j}$ are the local pose positions, velocities and local continuous joints rotations with $j$ denoting the number of joints; $ \mathbf{c}^f \in \mathbb{R}^{4} $ is the foot contact features calculated with the heel and toe joint velocities.

\section{Details of Evaluation Metrics}
Given the standard feature extrator provided by \cite{guo2022generating}, we can obtain the feature $f_{real}$ of the real motion, the feature $f_{gen}$ of the generated motion, and the feature $f_t$ of the corresponding text.

\noindent\textbf{FID.} The Fréchet Inception Distance (FID) is an evaluation metric used to measure the similarity between the generated motion distribution and the real motion distribution. Which can be calculated as:
\begin{equation}
    \textit{FID} = \|\mu_{\text{real}} - \mu_{\text{gen}}\|^2 + \text{Tr}(\Sigma_{\text{real}} + \Sigma_{\text{gen}} - 2(\Sigma_{\text{real}}\Sigma_{\text{gen}})^{1/2})
\end{equation}
where $\mu$ and $\Sigma$ represent means and variances of the corresponding feature distributions. $\text{Tr}(\cdot)$ denotes the trace of a matrix.
\begin{figure}[t] 
	\includegraphics[width=0.9\linewidth]{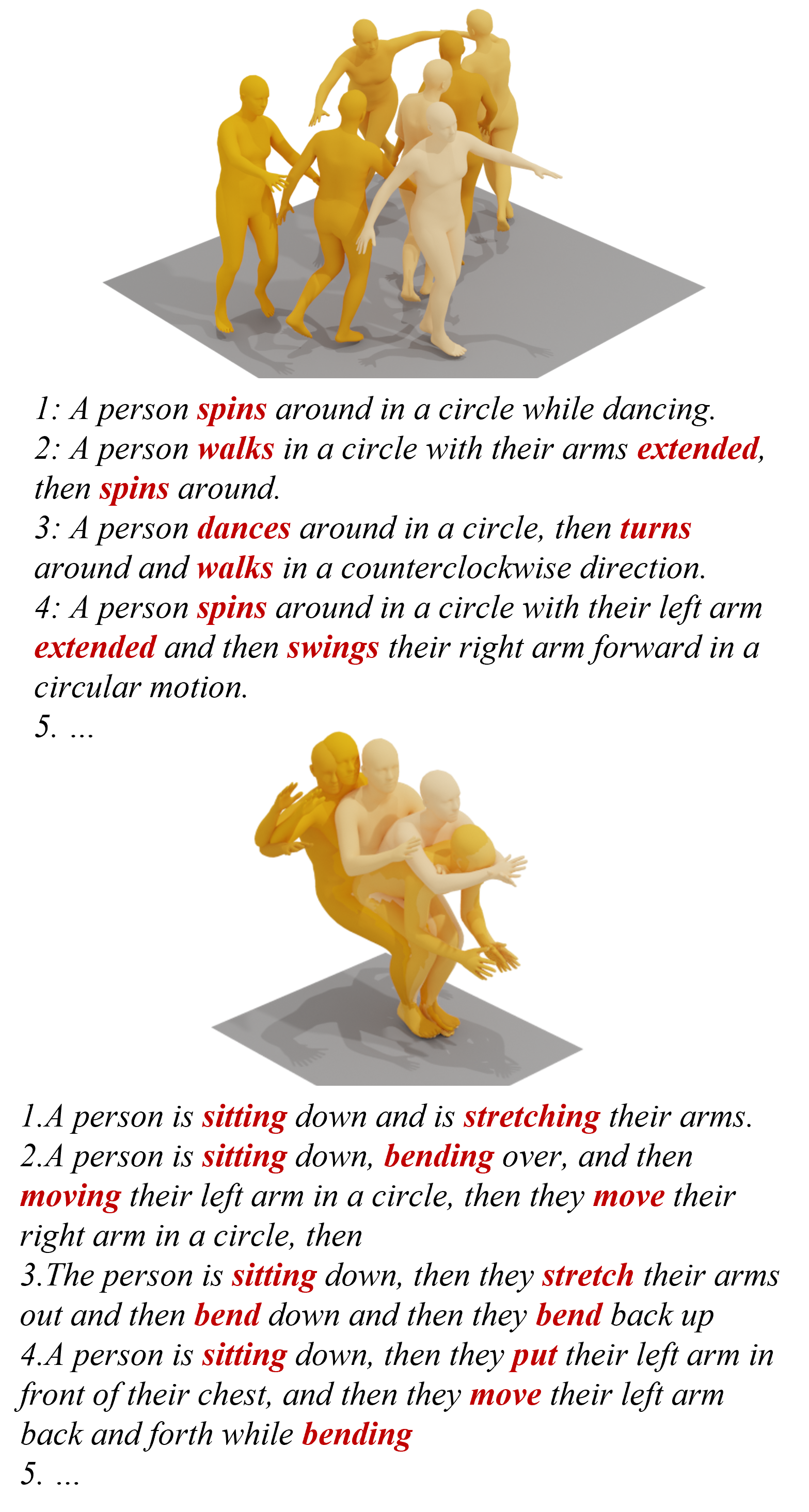} 
	\caption{Visualization of motion and diverse captions obtained from the WMC dataset.}
	\label{Fig:app_WMC} 
\end{figure}

\noindent\textbf{MM-Dist.} MM-Dist measures the similarity between motion features and text features. Given $N$ pairs of motion features $f_m$ and text features $f_t$, MM-dist can be calculated as:

\begin{equation}
    \textit{MM-Dist} = \frac{1}{N} \sum_{i=1}^{N}  \|f_m^i - f_t^i\|
\end{equation}

\noindent\textbf{Diversity.} Following \cite{guo2022generating}, We use the metric diversity to measure the variance of the whole motion features. Upon acquiring a set of 600 randomly sampled motion features and dividing them into two groups $\{f_m^i\}_{i=1}^{300}$ and $\{f_m^i\}_{i=301}^{600}$, Diversity can be calculated as:
\begin{equation}
    \textit{Diversity} = \frac{1}{300} \sum_{i=1}^{300}  \|f_m^i - f_m^{i+300}\|
\end{equation}

\noindent\textbf{MModality.} Following \cite{guo2022generating}, We use the metric MModality to measure the variance of motion features generated with the same text description. Given $N$ different text descriptions, we generate 30 motion samples for j-th text description and random sample two subsets $\mathcal{S}_{1,j}=\{f_{m,i,j}\}_{i=1}^{10}$ and $\mathcal{S}_{2,j}=\{f'_{m,i,j}\}_{i=1}^{10}$ containing 10 samples.
\begin{equation}
    \textit{MModality} = \frac{1}{10N} \sum_{j=1}^{N} \sum_{i=1}^{10}  \|f_{m,i,j} - f'_{m,i,j}\|
\end{equation}

\section{Qualitative Results.} 
\begin{figure*}[t] 
	\centering 
	\includegraphics[width=1\textwidth]{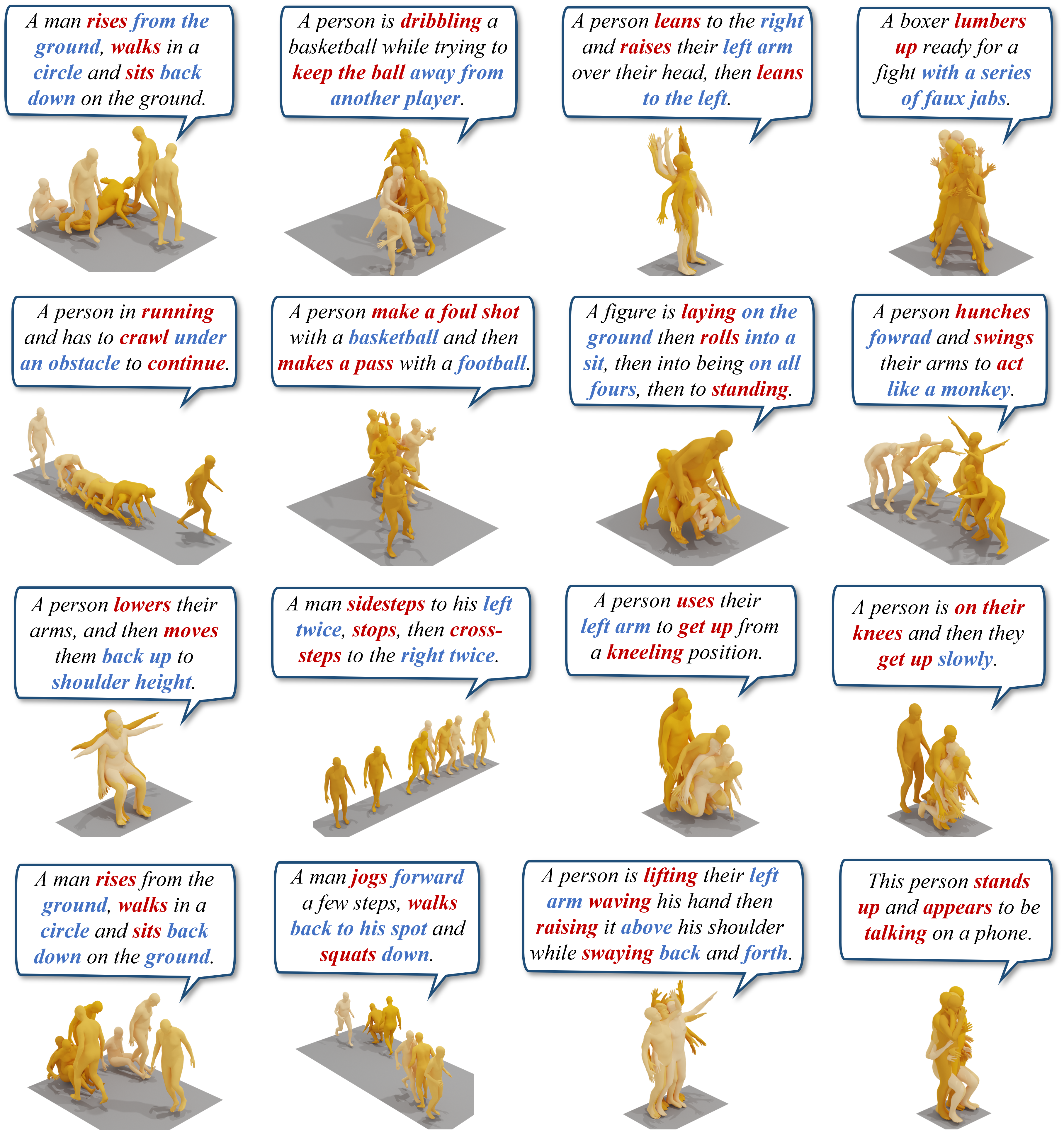} 
	\caption{More examples of visualizations of the generated results of our model with classifier-free scale $(s=2)$. The text of the motion descriptions is taken from the Humanml3d test set. we mark the actions that appear in the sentences in red, and the descriptions of the actions in blue.}
	\label{Fig:cp_vis1} 
\end{figure*}

\begin{figure*}[t] 
	\centering 
	\includegraphics[width=1\textwidth]{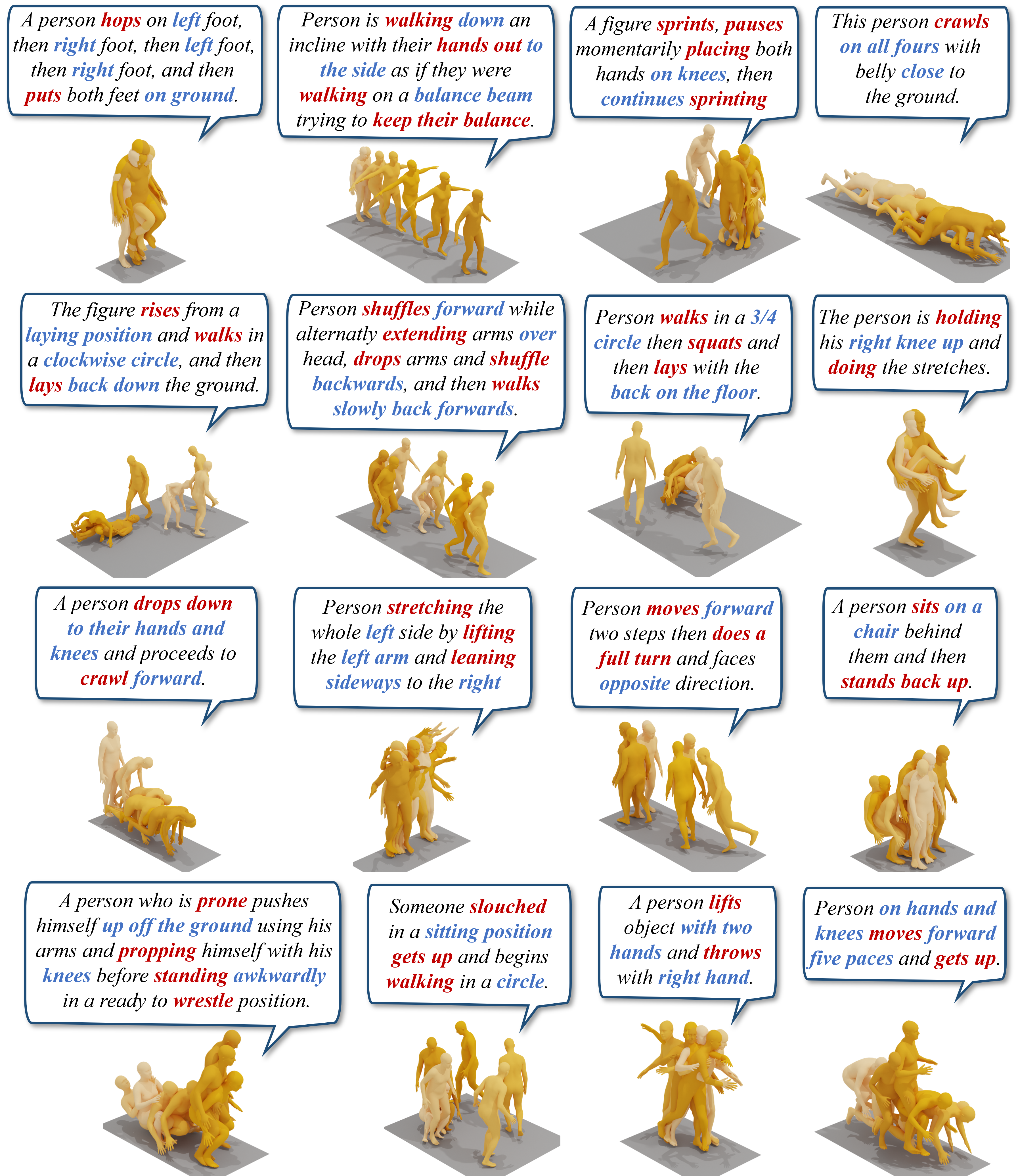} 
        \caption{More examples of visualizations of the generated results of our model with classifier-free scale $(s=2)$. The text of the motion descriptions is taken from the Humanml3d test set. we mark the actions that appear in the sentences in red, and the descriptions of the actions in blue.}
	\label{Fig:cp_vis2} 
\end{figure*}

Examples of motions and captions obtained in the WMC dataset are displayed in Figure ~\ref{Fig:app_WMC}. We show the visualization of our model on the text of the HumanML3D test set on Figure ~\ref{Fig:cp_vis1} and Figure ~\ref{Fig:cp_vis2}.



\end{document}